\documentclass[10pt,twocolumn,letterpaper]{article}
\usepackage[]{cvpr}      
\usepackage[pagebackref,breaklinks,colorlinks]{hyperref}
\usepackage{amsmath}
\usepackage[capitalize]{cleveref}
\crefname{section}{Sec.}{Secs.}
\Crefname{section}{Section}{Sections}
\Crefname{table}{Table}{Tables}
\crefname{table}{Tab.}{Tabs.}

\def\cvprPaperID{1} 
\def\confName{CVPR}
\def\confYear{2023}
\usepackage{url}            
\usepackage{booktabs}       
\usepackage{amsfonts}       
\usepackage{nicefrac}       
\usepackage{microtype}      
\usepackage{xcolor}         
\usepackage{caption}
\usepackage{subcaption}
\usepackage{algorithm2e}
\usepackage{blindtext}
\usepackage[utf8]{inputenc}
\usepackage{pgfplots}

\DeclareUnicodeCharacter{2212}{−}
\usepgfplotslibrary{groupplots,dateplot}
\usetikzlibrary{patterns,shapes.arrows}
\pgfplotsset{compat=newest}
\RestyleAlgo{ruled}

\title{SoccerNet 2023 Tracking Challenge - 3rd place MOT4MOT Team Technical Report}

\author{Gal Shitrit\thanks{Equal contribution}, \space Ishay Be'ery\footnotemark[1], \space  Ido Yerhushalmy
\\ Amazon Prime Video Sports\\
\texttt{{galshi, ishaybee, idoy}@amazon.com}}

\begin{document}

\maketitle

\begin{abstract}
The SoccerNet 2023 tracking challenge requires the detection and tracking of soccer players and the ball. In this work, we present our approach to tackle these tasks separately. We employ a state-of-the-art online multi-object tracker and a contemporary object detector for player tracking. To overcome the limitations of our online approach, we incorporate a post-processing stage using interpolation and appearance-free track merging. Additionally, an appearance-based track merging technique is used to handle the termination and creation of tracks far from the image boundaries. Ball tracking is formulated as single object detection, and a fine-tuned YOLOv8l detector with proprietary filtering improves the detection precision. Our method achieves 3rd place on the SoccerNet 2023 tracking challenge with a HOTA score of 66.27.
\end{abstract}

\section{Introduction}
The SoccerNet 2023 tracking challenge presents a unique and challenging task of detecting and tracking both the soccer players and the ball. Tracking multiple objects in a dynamic and fast-paced sport such as soccer is challenging. In this work, we present an approach for formulating player and ball tracking as separate tasks.

\section{Related Work}
Multi-Object Tracking (MOT) is the task of identifying and maintaining multiple object trajectories or tracks from a video stream.
For the online scenario, often applied to live video streams, predictions cannot rely on future frames. A common approach is Tracking-by-detection~\cite{wojke2017simple, bewley2016simple, du2023strongsort}, a two-step approach that first detects the objects in each frame of the video, and associates these detections over time to form trajectories (tracks). One way to create coherent trajectories is by applying a constant velocity Kalman Filter (KF)~\cite{bewley2016simple}. To better distinguish between objects, OC-SORT~\cite{cao2022observation} added a velocity consistency term, whereas DeepSORT~\cite{wojke2017simple} addressed the isuse by integrating additional appearance cues~\cite{wojke2017simple}. StrongSORT~\cite{du2023strongsort} improved the latter by replacing the appearance model with Bag of Tricks~\cite{luo2019bag} (BoT), and the addition of a Camera Motion Compensation (CMC). DeepOC-SORT~\cite{maggiolino2023deep} combined components from both StrongSORT and OC-SORT holistically to benefit from both approaches.
When information from future frames is available, for instance, when there is a delay between broadcasting and acquiring the video, additional offline methods may be applied. StrongSORT++~\cite{du2023strongsort} proposed to impute missing detections by interpolating and smoothing using a Gaussian process (GSI). Additionally, tracks are merged using the Appearance Free Linking (AFLink) model.

\begin{figure*}[t]
        \centering
        \includegraphics[page=1, trim = 14mm 160.2mm 170mm 12mm, clip,width=0.7\textwidth]{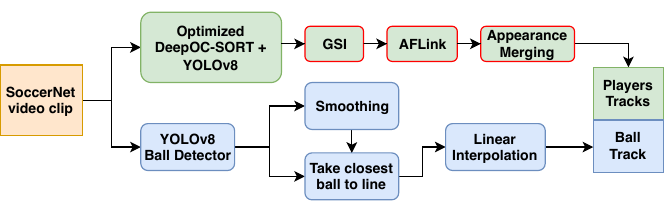}
        \caption{Tracking algorithm flow. The ball tracking components are marked in blue and the player tracking components are marked in green. The post-processing steps for the player tracker are marked with a red line.}
        \label{fig:design}

\end{figure*}

\section{Method}
In soccer games, there is a significant difference between ball and player tracking. Thus, the presence of multiple players in each frame contrasts with the single occurrence of a ball, and the ball can undergo considerable acceleration within short time intervals. Furthermore, ball detection is difficult due to its relatively small size, frequent occlusions, and tendency to blend in with the players' uniforms or crowd. Hence, our tracking approach was split to separate  ball and player tracking. The overall algorithm flow can be seen in Fig.~\ref{fig:design}.

\noindent
{\bf Player Tracking \quad}
Player tracking was approached through a two-step approach. In the initial step, an optimized state-of-the-art (SoTA) online multi-object tracker was employed in conjunction with a contemporary object detector. However, due to the inherent limitations of this online approach, unable to predict future events or modify past outcomes, a post-processing stage was applied to refine the tracking data. The post-processing comprises three phases. First, missing track detections are interpolated using GSI. Secondly, a fine-tuned AFLink model is utilized to merge tracks. The combination of these two techniques is denoted as "++". Last, an appearance-based track merging is employed assuming that, in our particular setup, new track IDs can only be created or terminated near the image boundaries. As a result, an iterative merging process is applied to identify track IDs that were terminated far from the image boundaries. Subsequently, if a new track is generated within a short time span, an attempt is made to merge it with the previously terminated track if their appearance similarity remains consistently high throughout most of the track duration.

\noindent
{\bf Ball Tracking \quad}
We approached ball tracking by treating it as a single object detection problem. In order to maximize the detection recall, we employed a fine-tuned YOLOv8l~\cite{JocherYOLObyUltralytics2023} detector with a low confidence score of 0.05. This configuration yields multiple detections for each frame within the video clip. To enhance the precision of the detections, we leveraged the temporal nature of the data through a series of filtering techniques. First, the center coordinates of the most confident detection in each frame were smoothed using 3rd order polynomial with a temporal window size of 51 frames. Subsequently, we retained only the detection closest to this smoothed trajectory if its distance was below 100 pixels. To address any missing detections, the resulting detections were then linearly interpolated to ensure tracking box continuity.

\section{Experiments}
We present a process aimed to improve and assess the performance of the MOT tracker by optimizing its individual components, namely the appearance model, detector, and tracker type, followed by their integration and evaluation. The optimization process involved fine-tuning each component in isolation and assessing its effectiveness. Subsequently, the optimized components were integrated into the tracker to evaluate the overall MOT performance.

\subsection{Data Preparation and Setup}
\noindent
{\bf Player Detector \quad}
Our tracker heavily depends on the performance of the player detector. To improve performance on the target domain, a YOLOv8l detection model was fine-tuned to detect players on SoccerNet~\cite{cioppa2022soccernet} data, and the Average Precision (AP) at IoU of 0.5 was measured. The image resolution was 576x1024, with a Non-Maxima Suppression (NMS) IoU threshold of 0.45.
The bounding boxes of players were extracted from 1710 frames of the tracking training set. These frames were selected by sampling at a rate of 1 frame per second (FPS) from all available training clips. Additionally, 200 manually inspected frames from the tracking test set were selected for validation. We note that the labels were not modified or changed in any manner during this process.

\noindent
{\bf Ball Detector \quad}
A separate YOLOv8l ball detection model was fine-tuned to detect the ball on SoccerNet data and the AP at IoU of 0.5 was measured. The image resolution was 1080x1920.
The bounding boxes of balls were extracted from a total of 2084 frames belonging to the tracking training set. Due to errors in the ground-truth labels, a COCO2017 pre-trained detector was used to select only frames with an IoU$>$0.7 between the label and the ball detection. This procedure ensures that only accurate labels appear in the training set. Additionally, 200 manually inspected frames from the tracking test set were selected for validation. The labels were not modified or changed in any manner during this process.

\noindent
{\bf Appearance Model \quad}
An appearance model (OSNet-ain~\cite{zhou2019omni}) was fine-tuned to match the target domain. Three different sizes of the model were trained, the smallest being $\times0.25$ and the largest $\times1$. The crop resolution was 256x128. The Rank-1 accuracy was measured along with the mAP.
The appearance dataset contained IDs extracted from the tracking train set. To ensure ID consistency between different clips, the team metadata, track ID, and jersey number metadata was utilized. Tracks without known jersey numbers were discarded from the dataset. The dataset comprises six distinct games, three games for the training set and three games for the validation set, featuring a total of 199 unique IDs, with 123 IDs in the training set and 76 IDs in the validation set. The training set encompasses an average of 1600 images per ID, while the validation set is composed of 10 images per ID, featuring eight galleries and two query images, totaling 760 images. The validation images were drawn from the tracking test set.

\noindent
{\bf Player Tracker \quad}
Two different trackers were evaluated with the fine-tuned components. The appearance model was used with cosine similarity. Furthermore, the upper limit of the tracker's performance was assessed by conducting the same experiment using GT boxes. The HOTA metric~\cite{luiten2021hota} was used for evaluation on SoccerNet test set.

\section{Results}

\noindent
{\bf Player Detector \quad}
Fine-tuning the player detector on the SoccerNet data improves its accuracy (AP@0.5 0.954 to 0.994), following Table~\ref{table:detection_perf}. After analyzing the failure cases, we found that some errors occur when several players overlap with each other.

\begin{table}[t]
  \caption{Performance of the player detection model on the test set when trained over different datasets.}
  \label{table:detection_perf}
  \centering
  \small
  \begin{tabular}{l c c c}
    \toprule
    Training Data  & AP@0.5  $\uparrow$ & AP@0.5:0.95  $\uparrow$\\
    \midrule
    COCO2017 & 0.954 & 0.812 \\
    SoccerNet & \bf{0.990} & \bf{0.923} \\
    \bottomrule
  \end{tabular}
\end{table}

\noindent
{\bf Ball Detector \quad}
Fine-tuning the ball detector results in AP@0.5 of 0.95 and AP@0.5:0.95 of 0.71. This indicates that the bounding box produced by the detector is not sufficiently tight. Furthermore, our investigation revealed that the detector struggles in scenarios involving partial occlusion, as well as when the ball's visual characteristics merge with other objects such as the crowd or white shoes.

\noindent
{\bf Appearance Model \quad}
Detection models of different sizes achieved similar mAP scores, with crop augmentation providing the greatest improvement, following  Table~\ref{table:reid}. The comparability of performance can be attributed to a multitude of factors: the relatively low number of identities (199 IDs as opposed to 1501 IDs in Market1501~\cite{zheng2015scalable}), which can lead to rapid overfitting, errors in training data that impede model improvement, and the nearly indistinguishable appearances of certain players on the same team.

\begin{table}[t]
  \caption{Appearance model performance for different model sizes and train augmentations.}
  \label{table:reid}
  \centering
  \begin{tabular}{l c c c}
    \toprule
    Model     & Augmentations   & Rank-1  $\uparrow$\ & mAP  $\uparrow$\\
    \midrule
    OSNet-x0.25  & None         & 0.83            & 0.79 \\
    OSNet-x0.25  & Crop        & 0.93            & 0.77 \\
    OSNet-x0.75  & Crop        & 0.95            & 0.8 \\
    OSNet-x0.75  & Crop, Flip  & \textbf{0.95}      & \textbf{0.8} \\
    OSNet-x1  & Crop        & 0.94           & 0.78 \\
    \bottomrule
  \end{tabular}
\end{table}

\begin{table}[t]
  \caption{Player tracking performance on SoccerNet test set for different trackers. Post-processing appearance merging is denoted by p.}
  \label{table:tracking}
  \centering
  \resizebox{0.47\textwidth}{!}{\begin{tabular}{lccc}
    \toprule
    Tracker     & Detector              & Configuration   & HOTA $\uparrow$\\
    \midrule
    Strong-SORT++ & YOLOv8       & OSNet-x1   & 65.18\\
    DeepOC-SORT++ & YOLOv8       & OSNet-x0.75   & 66.00\\
    DeepOC-SORT++p  & YOLOv8       & OSNet-x0.75   & \textbf{66.38}\\
    \midrule
    Strong-SORT++ & GT boxes      & OSNet-x1    & 85.26\\
    DeepOC-SORT++ & GT boxes      & OSNet-x0.75    & 87.85\\
    \bottomrule 
  \end{tabular}}
\end{table}

\begin{table}[t]
    \centering
  \caption{Ablation study of DeepOC-SORT's different components on SoccerNet train set: (post) merging post-processing, fine-tuned player detector (det), fine-tuned appearance model (app) and GSI + AFLink (++).}
  \label{table:trackingablation}
  \centering
  \begin{tabular}{ccccc}
    \toprule
     post & det &	app & ++ & HOTA $\uparrow$\\
    \midrule
      & & & & 57.3 \\
      & & & \checkmark & 59.0 \\
      & & \checkmark & \checkmark & 60.1 \\
     & \checkmark  & \checkmark & \checkmark & 66.0 \\
     \checkmark & \checkmark  & \checkmark & \checkmark & 66.4 \\
    \bottomrule 
  \end{tabular}
\end{table}

\noindent
{\bf Player Tracker \quad}
DeepOC-SORT++ achieves slightly better HOTA results than Strong-SORT++ (66.0\% and 65.2\% respectively), following Table~\ref{table:tracking}. The appearance merging during the post-processing further improves the HOTA by +0.38\%. Using GT boxes instead of detections greatly improves the HOTA metric by a notable margin of 21.85 points. The results suggest that the tracking accuracy is related to the detector's precision. This is further supported by the ablation study (see Table~\ref{table:trackingablation}), demonstrating that the fine-tuned detector has the most substantial impact on the HOTA metric, resulting in a gain of 5.9 points.

\noindent
{\bf Final Tracker \quad}
The final player and ball tracker achieved HOTA of 66.27, DetA of 70.32 and AssA of 62.62 on the challenge set.

\bibliographystyle{unsrt}
\bibliography{soccernet_report.bib}
\end{document}